\documentclass[sigconf, nonacm, pdfa]{acmart}

\newcommand\vldbdoi{XX.XX/XXX.XX}
\newcommand\vldbpages{XXX-XXX}
\newcommand\vldbvolume{14}
\newcommand\vldbissue{12}
\newcommand\vldbyear{2025}
\newcommand\vldbauthors{\authors}
\newcommand\vldbtitle{\shorttitle} 
\newcommand\vldbavailabilityurl{URL_TO_YOUR_ARTIFACTS}
\newcommand\vldbpagestyle{empty} 

\usepackage{enumitem}
\setlist[itemize]{noitemsep, left=0pt} % 针对 itemize 环境

\usepackage{xcolor} % 必需的颜色支持

\newcommand{\redcircle}[1]{%
  \textcolor{red!60!black}{\textcircled{\scriptsize #1}}%  % 70%红+30%黑
}

\newcommand{\greencircle}[1]{%
  \textcolor{green!50!black}{\textcircled{\scriptsize #1}}%  % 70%绿+30%黑
}          

\usepackage{subcaption}
\usepackage{stfloats}

\begin{document}
\title{ContextCache: Context-Aware Semantic Cache for Multi-Turn Queries in Large Language Models}

%%
%% The "author" command and its associated commands are used to define the authors and their affiliations.

\author{Jianxin Yan}
\affiliation{%
  \institution{Zhejiang University}
  \city{Hangzhou}
  \country{China}
}
\email{yanjianx666@gmail.com}

\author{Wangze Ni}
\authornote{Wangze Ni is also with The State Key Laboratory of Blockchain and Data Security; Hangzhou High-Tech Zone (Binjiang) Institute of Blockchain and Data Security.}
\affiliation{%
  \institution{Zhejiang University}
  \city{Hangzhou}
  \country{China}
}
\email{niwangze@zju.edu.cn}

\author{Lei Chen}
\affiliation{%
  \institution{HKUST (GZ) \& HKUST}
  \city{Guangzhou}
  \country{China}
}
\email{leichen@cse.ust.hk}

\author{Xuemin Lin}
\affiliation{%
  \institution{Shanghai Jiaotong University}
  \city{Shanghai}
  \country{China}
}
\email{xuemin.lin@gmail.com}

\author{Peng Cheng}
\affiliation{%
  \institution{Tongji University}
  \city{Shanghai}
  \country{China}
}

\email{cspcheng@tongji.edu.cn}

\author{Zhan Qin, Kui Ren}
\affiliation{%
  \institution{Zhejiang University}
  \city{Hangzhou}
  \country{China}
}
\email{{qinzhan,kuiren}@zju.edu.cn}

%%
%% The abstract is a short summary of the work to be presented in the
%% article.
\begin{abstract}

Semantic caching significantly reduces computational costs and improves efficiency by storing and reusing large language model (LLM) responses. However, existing systems rely primarily on matching individual queries, lacking awareness of multi-turn dialogue contexts, which leads to incorrect cache hits when similar queries appear in different conversational settings. This demonstration introduces ContextCache, a context-aware semantic caching system for multi-turn dialogues. ContextCache employs a two-stage retrieval architecture that first executes vector-based retrieval on the current query to identify potential matches and then integrates current and historical dialogue representations through self-attention mechanisms for precise contextual matching. Evaluation of real-world conversations shows that ContextCache improves precision and recall compared to existing methods. Additionally, cached responses exhibit approximately 10 times lower latency than direct LLM invocation, enabling significant computational cost reductions for LLM conversational applications.

\end{abstract}

\maketitle

%%% do not modify the following VLDB block %%
%%% VLDB block start %%%
\pagestyle{\vldbpagestyle}
\begingroup\small\noindent\raggedright\textbf{PVLDB Reference Format:}\\
\vldbauthors. \vldbtitle. PVLDB, \vldbvolume(\vldbissue): \vldbpages, \vldbyear.\\
\href{https://doi.org/\vldbdoi}{doi:\vldbdoi}
\endgroup
\begingroup
\renewcommand\thefootnote{}\footnote{\noindent
This work is licensed under the Creative Commons BY-NC-ND 4.0 International License. Visit \url{https://creativecommons.org/licenses/by-nc-nd/4.0/} to view a copy of this license. For any use beyond those covered by this license, obtain permission by emailing \href{mailto:info@vldb.org}{info@vldb.org}. Copyright is held by the owner/author(s). Publication rights licensed to the VLDB Endowment. \\
\raggedright Proceedings of the VLDB Endowment, Vol. \vldbvolume, No. \vldbissue\ %
ISSN 2150-8097. \\
\href{https://doi.org/\vldbdoi}{doi:\vldbdoi} \\
}\addtocounter{footnote}{-1}\endgroup
%%% VLDB block end %%%

%%% do not modify the following VLDB block %%
%%% VLDB block start %%%
\ifdefempty{\vldbavailabilityurl}{}{
\vspace{.3cm}
\begingroup\small\noindent\raggedright\textbf{PVLDB Artifact Availability:}\\
The source code, data, and/or other artifacts have been made available at \url{https://github.com/uYanJX/ContextCache}.
\endgroup
}
%%% VLDB block end %%%

\section{Introduction}

% Lorem ipsum dolor sit amet, consectetur adipiscing elit. Suspendisse a arcu quis arcu malesuada ultricies vitae in felis. Curabitur porta lacus at felis viverra hendrerit in non eros. Nam tempus tincidunt metus vitae fermentum. Donec sed risus felis. Cras luctus massa elementum, semper urna vel, efficitur ipsum. Morbi at tellus libero.

% Praesent imperdiet, lacus nec varius placerat, est ex eleifend justo, a vulputate leo massa consectetur nunc. Donec posuere in mi ut tempus. Pellentesque sem odio, faucibus non mi in, laoreet maximus arcu. In hac habitasse platea dictumst. Nunc euismod neque eu urna accumsan, vitae vehicula metus tincidunt. Maecenas congue tortor nec varius pellentesque. Pellentesque bibendum libero ac dignissim euismod. Aliquam justo ante, pretium vel mollis sed, consectetur accumsan nibh. Nulla sit amet sollicitudin est

\begin{figure}
  \centering
  \includegraphics[width=\linewidth]{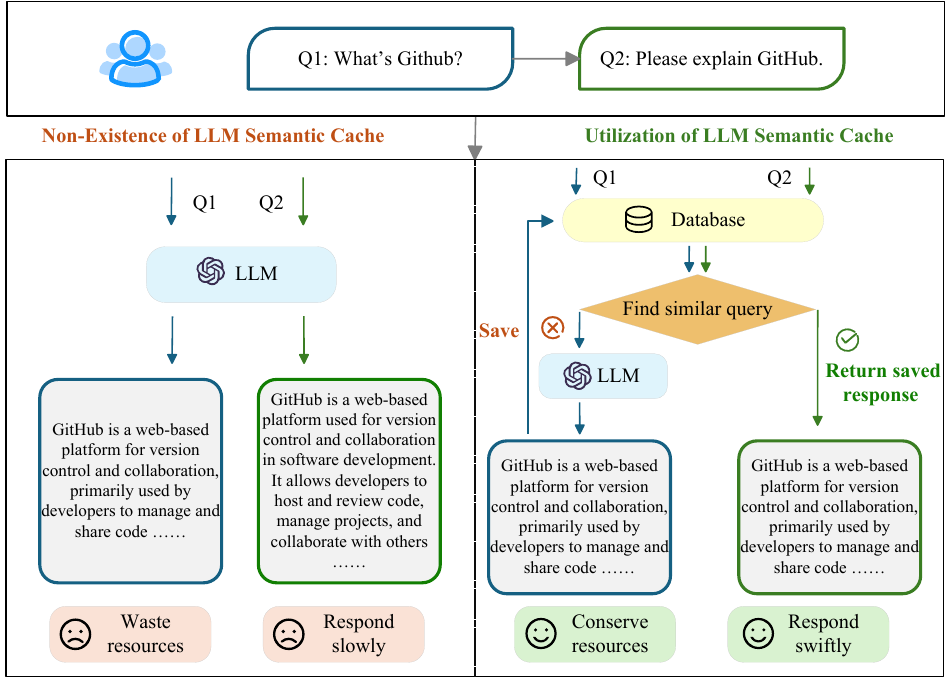}
  \caption{Optimizing LLM Responses with Semantic Caching}
  \label{fig:newconst}
  \vspace{-4mm}
\end{figure}

Large Language Models (LLMs) like ChatGPT have become essential components in conversational systems and productivity tools due to advanced language understanding and generation capabilities. However, LLM inference demands substantial computational resources, leading to significant operational costs. To address this challenge, semantic caching has emerged as an optimization strategy~\cite{bang2023gptcache}, adapting database result caching principles to LLM workflows through vector-based indexing and similarity-based retrieval of previously generated responses.

Figure~\ref{fig:newconst} illustrates LLM semantic caching, which enables the retrieval of previously computed responses for semantically similar queries. The figure contrasts two processing approaches: without caching (left), each query requires a complete LLM inference cycle, consuming significant computational resources and increasing latency; and with caching (right), when a user submits query Q2 that is semantically similar to a previously cached query Q1, the system directly retrieves the existing response without invoking the LLM. This cached approach \textbf{reduces both response time and operational costs by eliminating redundant processing}. Previous empirical studies support its practical efficiency: approximately 33\% of search engine queries are resubmitted~\cite{markatos2001caching}, and 31\% of ChatGPT interactions contain semantically similar queries~\cite{gill2024privacy}. 

Unlike traditional caching scenarios, \textbf{LLM interactions require consideration of dynamic conversational context rather than isolated requests}. Direct application of conventional database caching paradigms that match queries solely against stored embeddings achieves poor performance in LLM scenarios (as shown in Figure~\ref{fig:precision_recall}, where GPTCache~\cite{bang2023gptcache} exemplifies the conventional database caching paradigm). Existing approaches~\cite{gill2024privacy} attempt to concatenate previous queries before applying lightweight pre-trained models or average turn embeddings to extract contextual representations, but face two limitations: Concatenation-based methods encounter attention dilution when self-attention mechanisms process long text sequences~\cite{vaswani2017attention, tay2020long}, while embedding averaging causes representation flattening that obscures turn-specific semantic features~\cite{vaswani2017attention, reimers2019sentence}. Both approaches exhibit insufficient semantic discrimination due to limited sentence-level supervision, resulting in precision degradation as conversation length increases. Therefore, the core challenge for conversational caching systems is \textbf{to develop mechanisms that model context while maintaining precise semantic relationship detection with minimal computational overhead}.

To address these challenges, we present ContextCache, a context-aware semantic caching framework for multi-turn conversational systems. Our approach overcomes existing limitations through three key innovations. First, our Dynamic Context Modeling implements a \textbf{hierarchical self-attention mechanism} that captures cross-turn semantic dependencies, mitigating attention dilution in concatenation while preserving turn-specific features lost through representation flattening. Second, our LLM-enhanced training employs \textbf{difficult negative sample mining} to improve matching precision, addressing insufficient semantic discrimination through robust sentence-level supervision. Finally, our \textbf{Two-Stage Dynamic Retrieval Architecture} combines efficient vector retrieval for candidate selection with precise attention-based contextual matching, balancing computational efficiency with semantic accuracy. Experimental evaluation shows ContextCache outperforms GPTCache, \textbf{improving precision and recall by 10.9\% and 14.8\%} while cache-served responses deliver approximately \textbf{10 times lower latency} compared to LLM invocation. Our contributions include:

\begin{itemize}[topsep=0pt, itemsep=0pt, parsep=0pt]
    \item A Two-Stage Dynamic Retrieval Architecture combining vector retrieval with attention-based contextual matching.
    \item A prototype demonstrating our techniques in realistic conversational scenarios, reducing computational costs and latency.
\end{itemize}

\section{BACKGROUND}

\begin{figure}[t!]
   \vspace{-4ex}
    \centering
    \includegraphics[width=\linewidth]{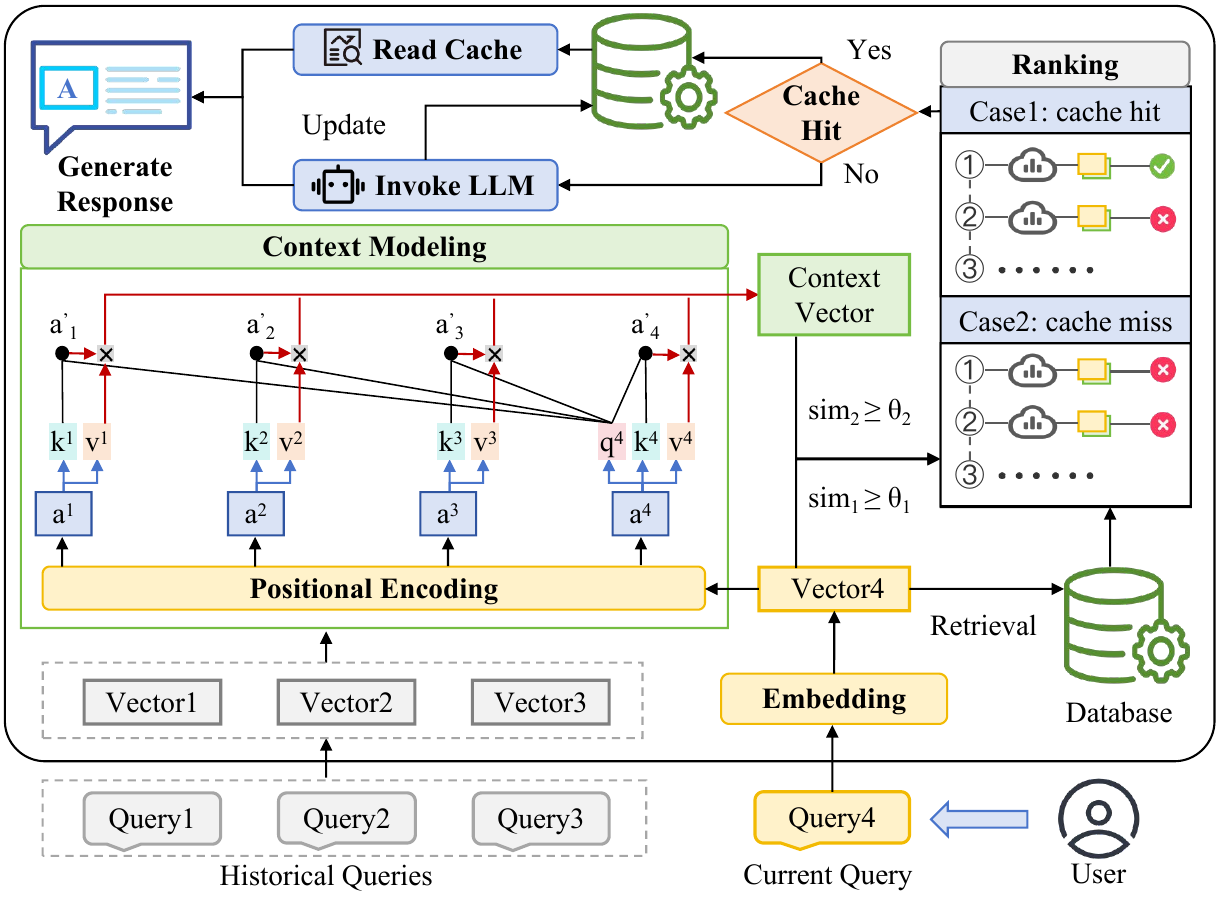}
    \caption{The Architecture Of ContextCache}
    \label{fig:sys}
    \vspace{-3mm}
  \end{figure}

Semantic caching improves upon traditional methods by operating on a query's meaning rather than its exact words. This allows the system to identify semantically equivalent questions, even when phrased differently, and reuse answers for greater efficiency.

\noindent{\textbf{Embedding generation}}. The core of the system is the embedding generator, which transforms queries into vector representations. It employs transformer-based architectures~\cite{vaswani2017attention} to encode semantic information: $E(q) = f_\theta(q)$, where geometric proximity in the vector space corresponds to semantic similarity~\cite{lan2019albert}.

\noindent{\textbf{Similarity quantification}}. This step provides the decision mechanism for the cache. It uses cosine similarity to measure the angular proximity between two embedding vectors, formulated as: $\text{similarity}(E_1, E_2) = \frac{E_1 \cdot E_2}{||E_1|| \, ||E_2||}$. When this similarity value exceeds a predetermined threshold, the system classifies queries as equivalent and retrieves cached response instead of invoking the LLM.

\section{SYSTEM OVERVIEW}

\begin{figure}
  \vspace{-4ex}
  \centering
  \begin{subfigure}[b]{0.35\textwidth}
    \centering
    \includegraphics[width=\textwidth]{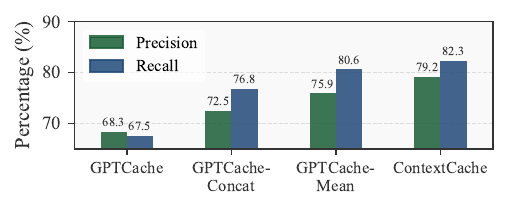}
    \caption{Precision and Recall}
    \label{fig:precision_recall}
  \end{subfigure}

 \vspace{0.01cm}
  
  \begin{subfigure}[b]{0.35\textwidth}
    \centering
    \includegraphics[width=\textwidth]{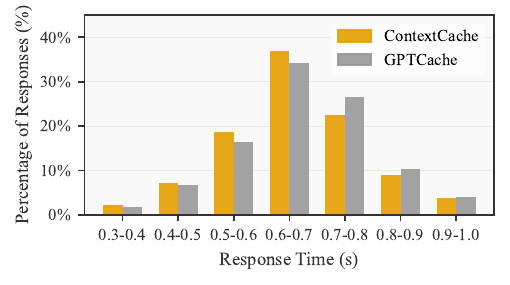}
    \caption{Cache Hit Response Time}
    \label{fig:confusion}
  \end{subfigure}
  \caption{Performance Metrics: ContextCache vs. GPTCache}
  \label{fig:performance}
  \vspace{-2mm}
\end{figure}

\subsection{Workflow}

ContextCache optimizes large language model applications by leveraging conversational context for semantic caching. Figure~\ref{fig:sys} illustrates the system architecture with six key components: (1) \textbf{Query interception}: User queries are intercepted before transmission to the LLM. (2) \textbf{Context collection}: The system captures the current query and retrieves historical dialogue to construct a comprehensive conversational context. (3) \textbf{Semantic representation generation}: The system generates embedding vectors only for the current query while reusing historical dialogue embeddings from previous calculations. (4) \textbf{Two-tier retrieval}: The system first employs the current query's embedding for preliminary similarity search to identify potential matches. For filtered candidates, it then integrates embeddings from both the current query and historical dialogue, applying self-attention mechanisms to analyze inter-turn relationships and generate a unified contextual representation for precise matching. (5) \textbf{Response determination}: When a cache entry with matching semantics and context is found, the system returns the cached response; otherwise, it forwards the query to the LLM. (6) \textbf{Cache update}: For LLM-generated responses, the system stores the response along with the query embedding and contextual representation before returning it to the user.

\begin{figure*}
  \vspace{-4ex}
  \centering
  \includegraphics[width=\textwidth]{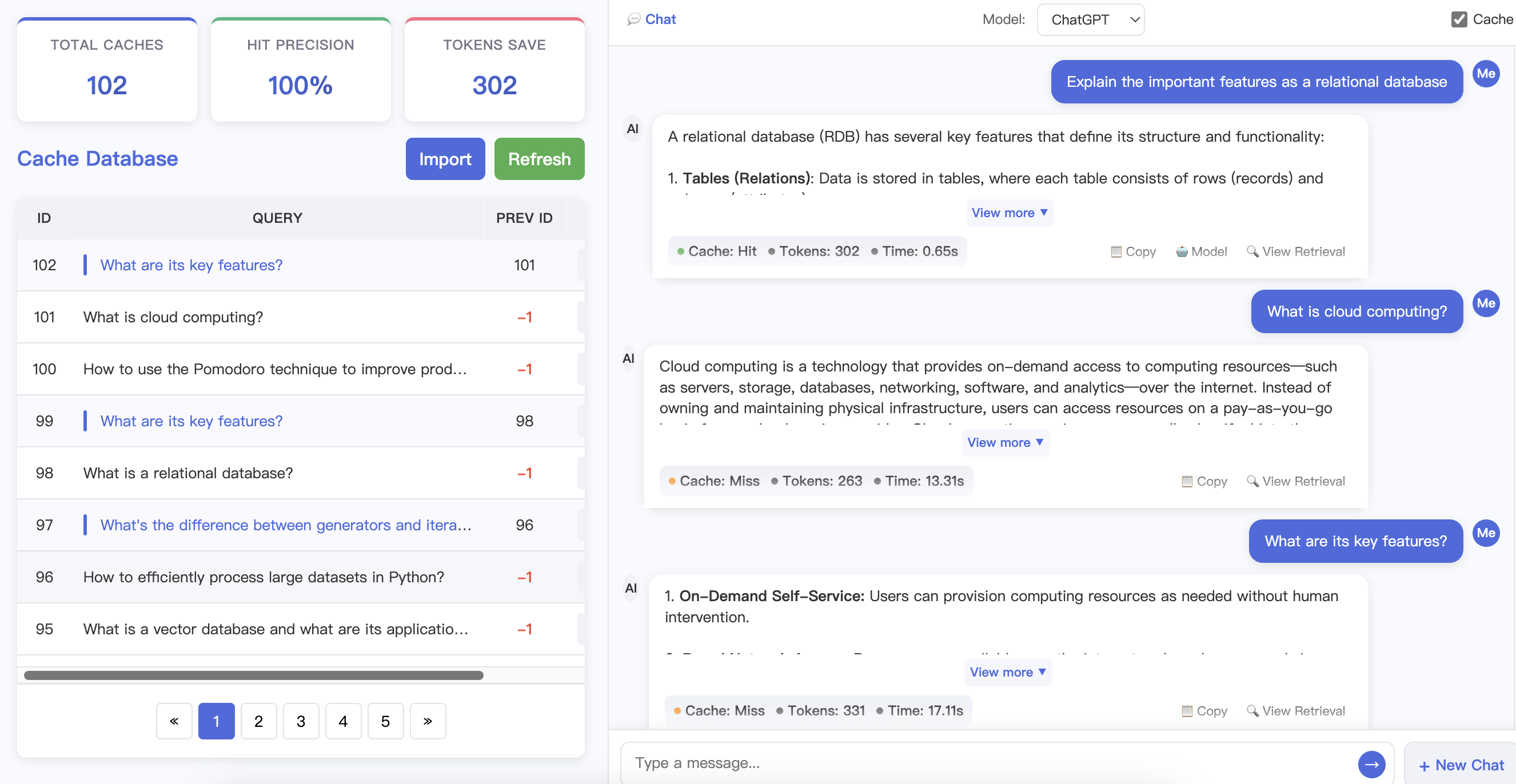} % 使用\textwidth自动适应双栏宽度
  \caption{ContextCache User Interface And Demonstration Scenarios}
  \label{fig:demo}
\end{figure*}

\subsection{Implementation} 
We enhanced GPTCache~\cite{bang2023gptcache} through an architecture that integrates conversational context: 
\begin{itemize}[topsep=0pt, itemsep=0pt, parsep=0pt]
    \item \textbf{Query preprocessing module} standardizes queries and incorporates conversation context. Historical dialogue records $H = \{h_1, h_2, ..., h_n\}$ establish the conversational context. The Albert~\cite{lan2019albert} embedding model $E$ generates semantic representation only for the current query $v_{Q} = E(Q)$, while historical dialogue vectors $V_H = \{E(h_1), E(h_2), ..., E(h_n)\}$ are reused from previous calculations, eliminating redundant computation.

    \item \textbf{Two-stage retrieval mechanism} improves cache hit accuracy through progressive refinement. The coarse-grained stage filters candidates $C_1 = \{c \mid \cos(v_{Q}, v_c) > \theta_1\}$ using cosine similarity threshold $\theta_1$, where $v_c$ represents the query vector of a cached entry. The fine-grained stage uses precomputed contextual representations stored in the cache. For each candidate, the global representation $g_c$ is retrieved from previous cache updates. The system generates the current conversation's global representation $g_{current} = \text{SelfAttention}(V_{current})$ where $V_{current} = \{v_{Q}\} \cup V_H$. This self-attention mechanism analyzes inter-turn relationships, producing a unified contextual representation. Final similarity assessment uses cosine similarity: $S_c = \cos(g_{current}, g_c)$. The system selects the optimal match $C_{best} = \arg\max_{c \in C_1} S_c$ that exceeds the threshold $\theta_2$, ensuring contextually appropriate responses while maintaining efficiency.

    \item \textbf{Cache update module} employs a dual-storage architecture. New responses $R$ and associated metadata are stored as contextual tuples $(Q, R)$ in a relational database, while query vectors $v_{Q}$ and global representations $g_{current}$ are indexed in the vector database for efficient similarity search. This separation optimizes both storage and retrieval performance. 
    
\end{itemize}

\subsection{Core Technique And Evaluation} 

Our evaluation used 1,000 queries derived from the ShareGPT dataset~\cite{sharegpt2024}. We initialized the cache with 30\% of original dialogue samples~\cite{markatos2001caching,gill2024privacy}, then generated the test queries by creating semantically equivalent variations through GPT-4-based paraphrasing to introduce linguistic diversity while preserving semantic intent. This methodology simulates real-world scenarios where users reformulate queries using different phrasings without altering the underlying conversational context.

\textbf{Context integration}. Our self-attention mechanism models semantic relationships across dialogue turns, generating unified contextual representations that capture inter-turn dependencies. Figure~\ref{fig:precision_recall} demonstrates that our approach significantly outperforms concatenation and averaging methods by preserving turn-specific semantic features while modeling their contextual relationships. Compared to GPTCache, this context-aware approach \textbf{achieves 10.9\% higher precision and 14.8\% improved recall}, particularly in distinguishing between semantically similar queries that appear in different conversational contexts.

\textbf{Efficiency optimization}. Our system reduces computational overhead through the two-stage retrieval mechanism and dual-storage architecture. Compared to direct LLM invocation, this design delivers cache-served responses with approximately \textbf{10 times lower latency}, substantially reducing operational costs (Figure~\ref{fig:dashboard}). Furthermore, as Figure~\ref{fig:confusion} demonstrates, despite incorporating contextual processing that improves semantic precision, we achieve a \textbf{3\% reduction in average cache hit time} compared to GPTCache.

\section{DEMONSTRATION OVERVIEW}

\begin{figure*}
   \vspace{-4ex}
    \centering
    \includegraphics[width=.85\textwidth]{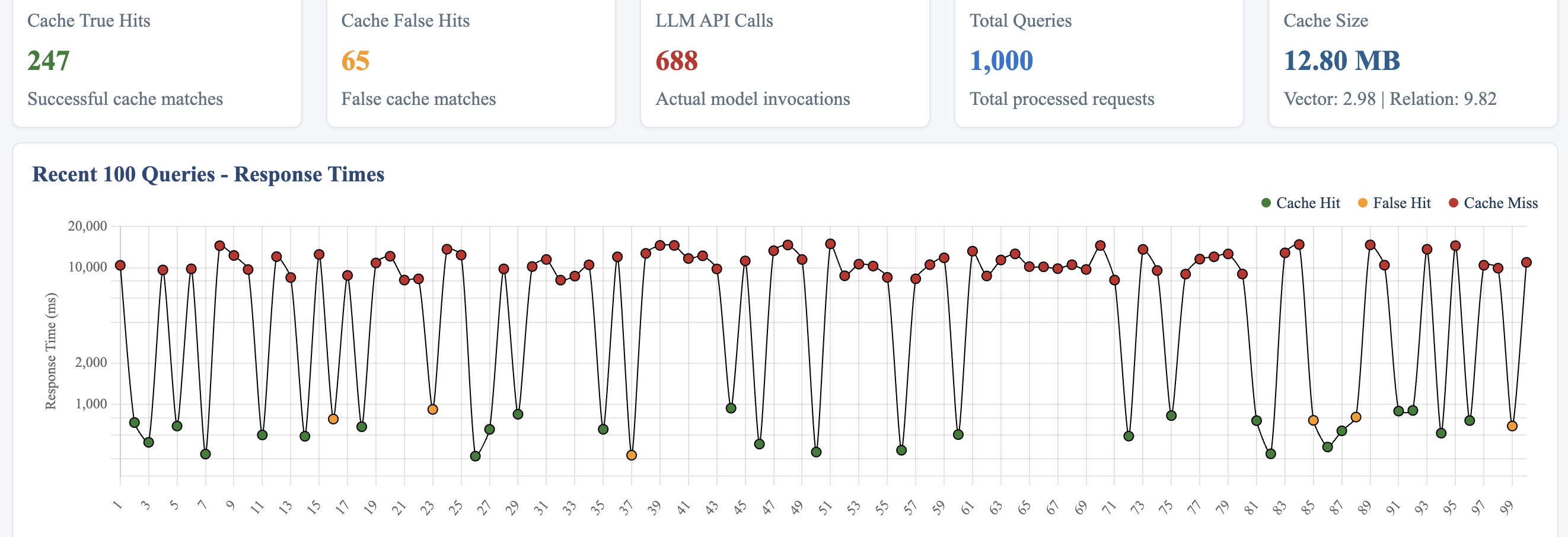} % 使用\textwidth自动适应双栏宽度
    \caption{Demonstration Engagement Results}
    \label{fig:dashboard}
\end{figure*}

We present ContextCache, a system featuring a Vue-based interface that extends LLM query caching through contextual awareness. Similar to GPTCache~\cite{bang2023gptcache}, our implementation uses Albert~\cite{lan2019albert} as the semantic encoder and employs a dual-database architecture combining SQL relational storage with FAISS vector indexing for retrieval. Our demonstration showcases key scenarios that illustrate the system's capabilities and practical benefits, as shown in Figure~\ref{fig:demo}.

\textbf{Scenario 1: Cache Hit}. \textit{Our first demonstration evaluates the system's ability to identify semantically similar queries despite significant linguistic variations, a critical capability for effective semantic caching in conversational systems}. 

The system features a split-screen layout with database contents and interaction panel. We populate it with a pre-curated collection of 100 query-response pairs containing "What is a relational database?" and "What are its key features?". For testing, users submit "Explain the important features as a relational database"—semantically equivalent but lexically distinct from the cached query. 

\greencircle{1} When the test query is submitted, the system retrieves the cached response, confirmed by a green "Cache Hit" indicator.  

\greencircle{2} Performance metrics display showing response time (0.65 seconds) and computational savings (302 tokens).

% This scenario demonstrates that context-aware semantic caching can effectively replace redundant LLM calls for common conversation patterns, reducing response latency to millisecond-level. %The system distinguishes between surface-level language differences and core semantic intent, demonstrating that context-enhanced embedding techniques can effectively replace redundant LLM calls for common conversation patterns.

This scenario illustrates how context-aware semantic caching effectively distinguishes between surface-level linguistic variations and contextual semantic intent. By successfully identifying semantic equivalence within specific conversational contexts, ContextCache enables a reduction in response latency from seconds to milliseconds while preserving response appropriateness.

\textbf{Scenario 2: Cache Miss}. \textit{Our second demonstration addresses a key challenge: correctly handling similar queries that require different responses based on their conversation history}. 

To demonstrate this capability, users initiate a conversation about cloud computing as a controlled test case. After entering "What is cloud computing?", they follow with "What are its key features?"—a query identical to the previously cached query about relational databases but requiring a context-appropriate response.

\redcircle{1} When this query is submitted, the system recognizes the contextual difference and displays a "Cache Miss" indicator.

\redcircle{2} Following response generation, the database panel updates to display the newly stored conversation with its preserved context.

This scenario highlights ContextCache's ability to prevent semantic confusion through effective contextual discrimination. By incorporating conversational history into the matching process, the system successfully distinguishes between similar queries that occur in different contextual settings, reducing false positive matches.

% \textbf{Demonstration engagement}. To quantify system effectiveness, we integrated an interactive dashboard (Figure~\ref{fig:dashboard}) that visualizes ContextCache's operational metrics in real-time. The interface displays comprehensive performance statistics from evaluations using 1,000 ShareGPT dataset~\cite{sharegpt2024} queries (with 30\% repeated queries preloaded as cached entries). Key metrics include 238 Cache Positive Hits, 48 Cache Negative Hits, 714 LLM API Calls, 1,000 Total Queries processed, and a 12.80 MB Cache Size. The dashboard's execution log reveals that cache-served responses deliver approximately 10 times lower latency compared to LLM invocation, enabling users to verify ContextCache's efficiency advantages while confirming high semantic matching precision in multi-turn dialogues.

\textbf{Demonstration engagement}. To quantify system effectiveness, we integrated an interactive dashboard (Figure~\ref{fig:dashboard}) that visualizes ContextCache's operational metrics in real-time. The interface tracks key performance indicators, including Cache True/False Hits, LLM API Calls, Query Volume, and Memory Utilization. The execution log provides request-level performance data, enabling users to examine query processing paths and response timing. Through this log analysis, users can clearly observe that cache-served responses deliver approximately \textbf{10 times lower latency} compared to LLM invocation, while maintaining high semantic matching precision in multi-turn dialogues. This demonstrates that ContextCache is an efficient solution for reducing computational costs and response latency in LLM applications.

\section{CONCLUSION}

Our work introduces ContextCache, a context-aware semantic caching system that reduces LLM inference costs, contributing to the wider adoption of LLM applications.
% \input{6ack}

% \begin{acks}
%  This work was supported by the [...] Research Fund of [...] (Number [...]). Additional funding was provided by [...] and [...]. We also thank [...] for contributing [...].
% \end{acks}

%\clearpage

\bibliographystyle{ACM-Reference-Format}
\bibliography{sample}

\end{document}